\documentclass{article}
\usepackage{ijcai21}
\usepackage{times}
\usepackage{soul}
\usepackage{url}
\usepackage[hidelinks]{hyperref}
\usepackage[utf8]{inputenc}
\usepackage[small]{caption}
\usepackage{graphicx}
\usepackage{amsmath}
\usepackage{amsthm}
\usepackage{booktabs}
\usepackage{algorithm}
\usepackage{algorithmic}
\usepackage{amsfonts}
\title{Trajectory Prediction for Autonomous Driving Using a Transformer Network}

\author{
Zhenning Li$^1$
\and Hao Yu$^2$
\affiliations
$^1$State Key Laboratory of Internet of Things for Smart City, University of Macau\\
$^2$School of Transportation, Southeast University \\
\emails
zhenningli@um.edu.mo, seudarwin@gmail.com}

\begin{document}

\maketitle

\begin{abstract}
Predicting the trajectories of surrounding agents is still considered one of the most challenging tasks for autonomous driving. In this paper, we introduce a multi-modal trajectory prediction framework based on the transformer network. The semantic maps of each agent are used as inputs to convolutional networks to automatically derive relevant contextual information. A novel auxiliary loss that penalizes unfeasible off-road predictions is also proposed in this study. Experiments on the Lyft l5kit dataset show that the proposed model achieves state-of-the-art performance, substantially improving the accuracy and feasibility of the prediction outcomes.


\end{abstract}

\section{Introduction}
Predicting the future trajectories of surrounding agents is critical for autonomous driving. This task is inherently challenging due to the significant variants in agents' preferences and actions.  However, in most cases, the future motion of one agent is closely related to its past states as well as the structure of the current scene. With this in mind, we believe that reasoning about contextual information of the scene can help a prediction model generate more plausible outcomes. On the other hand, it also implies that generating a finite set of hypothetical trajectories can describe the future motion of an agent to a large extent. Therefore, ideal predictions should be conditioned on the latent representation of the contextual features and can represent multiple possibilities and their associated likelihoods. Furthermore, we expect that predicted trajectories are plausible and conform to real-world constraints.


In the literature, data-driven models are the most successful category of approaches for our problem. One of the key components that ubiquitous in these models is the recurrent neural networks, especially the Long Short\textendash Memory (LSTM) architecture.  LSTMs can implicitly model the inherent dependence between successive observations of agents' trajectories in an end-to-end fashion. However, recent studies have raised criticisms of the memory mechanism of the LSTM and its ability of modeling longer sequential data ~\cite{bai2018empirical}.
Thus, Transformer networks were introduced and quickly became the preferred model when it comes to sequential modeling tasks such as natural language processing ~\cite{vaswani2017attention}.
This outstanding performance can be primarily contributed to its parallel computing framework as well as the usage of the multi-head attention mechanism.

%
%

To this end, we propose an approach based on Transformer network, namely Context-Aware Transformer (CATF), which takes both contextual information of the scene and historical states of the agent as inputs. The proposed research addresses above-mentioned issues in the following aspects. (1) The contextual features of the studied scene are extracted with a context-aware module by injecting the bird's eye view image of the scene. (2) A multi-modal trajectory prediction framework is carefully designed, which can obtain multiple hypothetical outputs with corresponding credibility. (3) A novel off-road loss is introduced, which is able to constrain the output to be more physically feasible. This loss together with the classification loss are combined in a multitask learning fashion. (4) In addition, we also introduce a CATF$_{l}$ model which adopts a linear projection technique to further reduce the computational burden of the multi-head attention module. To the best of our knowledge, this is one of the first paper using such framework for the trajectory prediction task. Experiments on l5kit dataset demonstrated the proposed CATF and CATF$_{l}$ have top-tier performance over baselines, results also showed that introducing the linear projection can significantly boost the inference time and save the memory usage while barely degrades the prediction accuracy.

\section{Problem Formulation}
Let us denote the overall map data as $\mathcal{M}$, and a set of $K$ discrete time steps at which the tracking system outputs state estimations as $\mathcal{T}_{K}$. The time gap between two consecutive time steps in $\mathcal{T}_{K}$ is set to be constant and corresponding to the reporting frequency (for example, when the tracking system running at a frequency of $10Hz$, the time gap between $t_{j-1}$ and $t_{j}$ equals $0.1s$). The state of $i$\textendash th agent ($ i \in \mathcal{N}_{t_j}$) at time $t_j$ is denoted as $\mathbf{s}_{i,t_j}$, where $\mathcal{N}_{t_j}$ is the set of unique agents tracked by the AV at $t_j$. It is noteworthy that due to the limitation of the sensing range,  the equality of the member of agents at different reporting times, i.e., $\mathcal{N}_{t_{j}}=\mathcal{N}_{t_{j^{'}}}, \forall j\neq j^{'}$, does not necessarily hold. For convenience, we assume we only focus on the agents appeared in the AV's perception region at the reference time $t_{j}$.  Correspondingly, the discrete-time trajectory of agent $i$ between $t_{j}$ and $t_{j+k}$ can be written as $\mathbf{s}_{i,t_{j}:t_{j+k}}=\{\mathbf{s}_{i,t_{j}},\mathbf{s}_{i,t_{j+1}},\ldots,\mathbf{s}_{i,t_{j+k-1}}\}$. Past and future states are represented in an actor-centered coordinate system derived from agent's state at time $t_j$, where forward direction defines $x\textendash$axis, left-hand direction defines $y\textendash$axis. Without loss of generality, in this study, we assume the state of the agent $i$ at $t_j$ can be described by its $x\textendash$ and $y\textendash$positions, that is, $\mathbf{s}_{i,t_{j}}=\left\lbrace x_{i,t_{j}},y_{i,t_{j}} \right\rbrace$. Other elements in the agent's state can be further deduced from the position information; for example, the headings and velocities can be inferred from the displacement vectors between consecutive positions.

Let $\mathcal{I}_{i}=\{\bigcup \mathbf{s}_{i,t_{j-m}:t_{j+1}}; \mathcal{M}\}$ denote the scene context over the past $m$ steps (i.e., partial historical and current states of the agent and the map data) at the given prediction reference time $t_{j}$. Therefore, our task is to predict future states $\mathcal{\hat{S}}_{i}=\{\mathbf{\hat{s}}_{i,t_{j+1}:t_{j+H+1}}\}, \forall i\in \mathcal{N}_{t_j}$, where $H$ is the prediction horizon. Note that this is an unimodal prediction task since only one possible trajectory for the agent $i$ is predicted.This task can be readily extended to a multi-modal task, which generates multiple hypothetical trajectories (up to $K$), $\mathcal{\hat{S}}_{i}^{K}=\{\mathbf{\hat{s}}^{k}_{i,t_{j+1}:t_{j+H+1}}, \forall k \in K \}$, further described by a credibility vector $\mathcal{C}_{i}^{K}=\{{c}_{i}^{1},\ldots,{c}_{i}^{K}\}$, and $\sum_{i=1,\ldots,K} {c}_{i}^{1} +  {c}_{i}^{2} +\ldots +{c}_{i}^{K} =1$.

\section{Context-Aware Transformer Model}
\begin{figure}[h]
	\centering
	\includegraphics[width=1\linewidth]{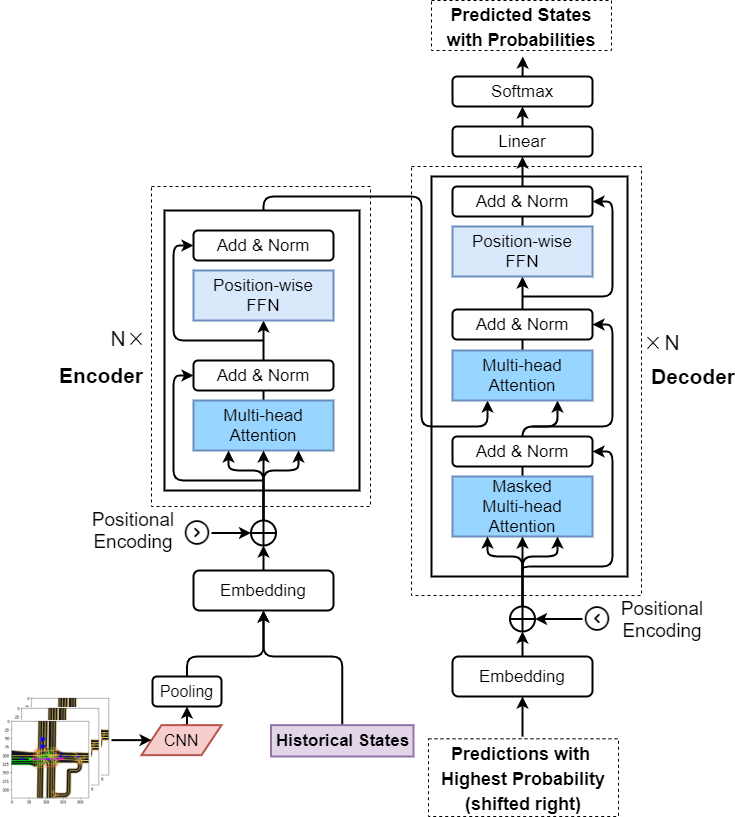}
	\caption{Structure of Transformer Network}
	\label{fig:st-attention-1}
\end{figure}

We propose a Context-Aware Transformer (CATF) model where each agent is modeled by a transformer network instance. Each transformer network predicts the future trajectory of the agent based on its previous states. The context information of the studying scene is injected to the model with the help of a CNN. We believe that by including such contextual information in addition to the location information, it can help improve the performance of the model. The overall framework is illustrated in Fig. \ref{fig:st-attention-1}. The framework is composed of several blocks, including Attention module, Position-wise Feed-Forward Networks module, and residual connection and layer normalization module after each of the previous blocks. The CATF network can capture the dependence of time series data and the nonlinear characteristics of spatial data through the multi-head attention module, which includes a self-attention mechanism and a multi-head attention mechanism. The following sections describe the implementation details.

\subsection{Encoder-Decoder Framework}
\subsubsection{A. Input and Output}
The Encoder-Decoder structure is extensively used in sequence transduction models ~\cite{sutskever2014sequence}. In this work, for the prediction task of agent $i$, the encoder network maps each element in the embedded input sequence to the latent state and the decoder network generates one element of the output sequence on every time step. The previously generated element is used as an additional input to generate the next element in an auto-regressive way. The scene map is first rasterized as an RGB image ($H\times W\times 3$) and serves as the input to a CNN. Note that any CNN architectures can be used as the backbone. A global average pooling layer is then added followed the CNN. The outputs and the historical states of both target agents and the AV are concatenated together and then embedded onto the higher \textit{D}\textendash dimensional space $\mathbf{e}_{i,t}, t \in (j-m,j+1)$ by a fully connected layer. In the same way, the output is also expanded into the \textit{D}\textendash dimension.

\subsubsection{B. Positional Encoding}
In order to use the sequence order information, we inject positional information by adding a fixed positional encoding based on sine and cosine functions to the input representations. The positional encoding outputs $\mathbf{e}_{i,t} + \mathbf{p}^{t}$ using a positional embedding matrix $\mathbf{p}^{t}=\{p_{t,d}\}_{d=1}^{D}$, whose element on the odd or the even column is
\begin{equation}\label{eq1}
	p_{t,d}=\left\{\begin{matrix}
		sin(\frac{t}{10000^{d/D}})& \text{for \textit{d} even}\\
		cos(\frac{t}{10000^{d/D}})& \text{for \textit{d} odd}
	\end{matrix}\right.
\end{equation}
In other words, each dimension of the positional encoding  changes in time according to sine waves of different frequencies, from $ 2 \pi $ to $ 10000 \cdot \pi $. This ensures that a sequence of up to 10,000 elements has a unique timestamp and extends the unseen length of the sequence.

\subsubsection{C. Multi-head Attention}
\begin{figure*}[h]
	\centering
	\includegraphics[width=0.7\linewidth]{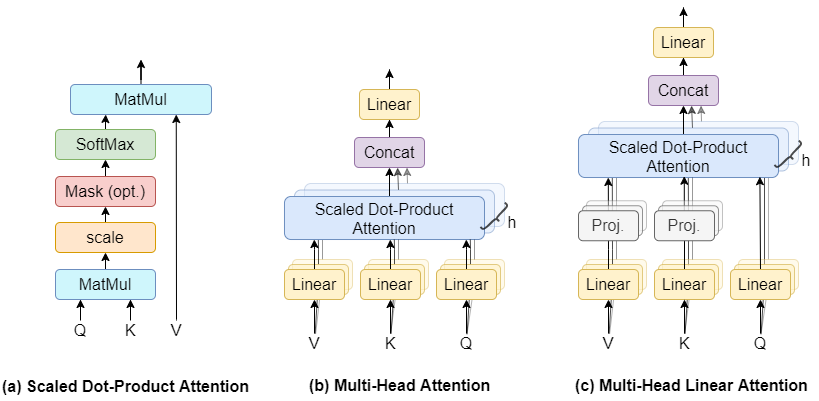}
	\caption{Frameworks of Scaled Dot-Product Attention, Multi-Head Attention, and Multi-Head Linear Attention}
	\label{fig:mutiattention}
\end{figure*}

Given a \textit{query} and a set of \textit{key-value} pairs, attention can be generalized to compute a weighted sum of the values dependent on the query and the corresponding keys. In other words, it selects the relevant content of the input to calculate each position represented by the output. Compared with CNNs or RNNs, the framework directly expands the receptive field. We assume the dimension of query and key vectors is $d_k$, and the dimension of value vectors is $d_v$. As shown in Fig. \ref{fig:mutiattention}(a), the scaled dot-product attention itself is computed by taking the dot-product attention between the query and key vectors divided by the square root of $d_k$ (i.e., the scaled operation) before finally passing them to the softmax function to get their weights by their values. The scaled operation is used to avoid the softmax function falling into regions where it has extremely small gradients.

Instead of performing a single attention pooling, $Q$, $K$ and $V$ can be transformed with $h$ times independently learned linear projections. Then these $h$ projected $Q$, $K$ and $V$ are fed into attention pooling in parallel. In the end, $h$ attention pooling outputs are concatenated and transformed with another learned linear projection to produce the final output. This design is called multi-head attention (as shown in Fig. \ref{fig:mutiattention}(b)), where each of the $h$ attention pooling outputs is a head. Benefiting from this multi-head attention operation, the transformer structure can collectively generate comprehensive latent features from trajectory data from different representation subspaces. Thus,
\begin{equation}\label{Eq2}
	\text{MultiHead}(Q,K,V)=\text{Concat}[\text{head}_{1},\ldots,\text{head}_{h}]W^{O}
\end{equation}
Each head is defined as:
\begin{equation}\label{Eq3}
	\begin{aligned}
		\text{head}_{i}&=\text{Attention}(QW_i^{Q}, KW_{i}^{K}, VW_{i}^{V})\\
		&=\underbrace{\text{softmax}\left[(\dfrac{QW_{i}^{Q}(KW_{i}^{K})^{T}}{\sqrt{d_k}})\right]}_{P}VW_{i}^{V}
	\end{aligned}
\end{equation}
where above $W$ are all learnable parameter matrices, and the Transformer uses $P$ to capture the input context. Obviously, computing $P$ is expensive as it requires multiplying two $n\times d_{k}$ matrices ($n=D$), which is $O(n^2)$ in time and space complexity. Recent works have argued that this quadratic dependency on the sequence length has become a bottleneck for Transformers ~\cite{wang2020linformer,kitaev2020reformer}.

Follow the work of ~\cite{wang2020linformer}, we used a linear attention mechanism (as shown in Fig. \ref{fig:mutiattention}(c)) which allows us to compute the contextual mapping $P\cdot VW_{i}^{V}$ in linear time and memory complexity with respect to the sequence length. The main idea of the linear attention is to add two linear projection (Proj.) matrices $F_{i},G_{i}\in \mathbb{R}^{n\times p}$ when computing key and value. The orignal ($n\times d_{k}$)\textendash dimensional key and value layers projected into ($p\times d_{k}$)\textendash dimensional key and value layers. The modified head $\text{head}_{i}^{'}$ can be calculated by multiplying contextual mapping matrix $P^{'} \in \mathbb{R}^{n\times p}$ and $G_{i}VW_{i}^{V}$, that is
\begin{equation}\label{Eq4}
	\begin{aligned}
		\text{head}_{i}^{'}&=\text{Attention}(QW_i^{Q}, F_{i}KW_{i}^{K}, G{i}VW_{i}^{V})\\
		&=\underbrace{\text{softmax}\left[(\dfrac{QW_{i}^{Q}(F_{i}KW_{i}^{K})^{T}}{\sqrt{d_k}})\right]}_{P^{'}:n\times p}\underbrace{G_{i}VW_{i}^{V}}_{p \times d_{v}}
	\end{aligned}
\end{equation}
Note that the above operations only require $O(np)$ time and space complexity. Obviously, if the projected dimension $p$ is set to be very small, such that $p\ll n$, then the computational burden of memory and space can be significantly reduced. This model can be further simplified by using only one learnable projection matrix $F_i$=$G_i$=$G$ across all layers for all heads and for both key and value. We refer the model uses such head as $\text{CATF}_{l}$.
\subsubsection{D. FFN and Add \& Norm Layers}
In addition to the attention layer, each Transformer layer also contains a fully connected feed-forward network, which is applied to each position separately and identically (i.e., position-wise). The feed-forward network is composed of two linear transformations with a sigmoid activation function in between. The following Add \& Norm module uses a residual connection to prevent the transformer from degrading due to multi-layer stacking (similar to the residual operation in the ResNet), and adopts a layer normalization to prevent the range of values in the layers from changing too much, which allows faster training and better generalization ability.
\subsection{Loss Functions}
Assume the ground truth positions of a sample trajectory of agent $i$ in the perdition horizon (from $t_{j+1}$ to $t_{j+H}$) are $(x_{i,t_{j+1}:t_{j+H+1}}, y_{i,t_{j+1}:t_{j+H+1}})=\left\lbrace (x_{i,t_{j+1}}, y_{i,t_{j+1}}),\ldots, (x_{i,t_{j+H}}, y_{t_{i,j+H}})\right\rbrace$, and we predict $K$ hypotheses, represented by $(\hat{x}_{i,t_{j+1}:t_{j+H+1}}^{k},\hat{y}_{i,t_{j+1}:t_{j+H+1}}^{k})=\left\lbrace (\hat{x}_{i,t_{j+1}}^{k}, \hat{y}_{i,t_{j+1}}^{k}),\ldots, (\hat{x}_{i,t_{j+H}}^{k}, \hat{y}_{i,t_{j+H}}^{k})\right\rbrace$. In addition, we also predict confidences $\mathcal{C}_{i}^{K}$ of these $K$ hypotheses. For convenience, the subscripts $i$ and are omitted in the following content of this section. In addition, we use $1:T$ to represent the period $t_{j+1}:t_{j+H+1}$, and let the sample element in this set be $t$. We assume the ground truth positions to be modeled by a mixture of multi-dimensional independent Normal distributions over time, yielding the likelihood

\begin{align}\label{eq5}
	\begin{split}
		& p\left( (x_{1:T}, y_{1:T})|
		c^{k}, (\hat{x}_{1:T}^{k}, \hat{y}_{1:T}^{k})\right) _{\forall k\in K}\\
		& =\sum_{\forall k}c^{k}\mathcal{N}\left (x_{1:T}|\hat{x}_{1:T}^{k}, \Sigma=1\right) \mathcal{N}\left(y_{1:T}|\hat{y}_{1:T}^{k}, \Sigma=1 \right )\\
		& =\sum_{\forall k}c^{k}\prod_{\forall t} \mathcal{N} \left (x_{t}|\hat{x}_{t}^{k}, \sigma=1\right)  \mathcal{N} \left (y_{t}|\hat{y}_{t}^{k}, \sigma=1 \right)
	\end{split}
\end{align}
We then use the negative log-likelihood of the ground truth data given these multi-modal predictions to calculate the classification loss $L_{c}$, that is,
\begin{align}\label{eq6}
	\begin{split}
		&  L_{c}=-\log\left ( p\left( (x_{1:T}, y_{1:T})|
		c^{k}, (\hat{x}_{1:T}^{k}, \hat{y}_{1:T}^{k})\right)_{\forall k\in K} \right)\\
		& = -\log\sum_{\forall k}e^{\log(c^{k}) + \sum_{\forall t}{\mathcal{N} \left (x_{t}|\hat{x}_{t}^{k}, \sigma=1\right)  \mathcal{N} \left (y_{t}|\hat{y}_{t}^{k}, \sigma=1 \right)}}\\
		& = -\log\sum_{\forall k}e^{\log(c^{k}) -\frac{1}{2}\sum_{\forall t}\left ( (x_{t}-\hat{x}_{t}^{k})^{2}+ (y_{t}-\hat{y}_{t}^{k})^{2}\right)}\\
		& = -a^{*}-\log\sum_{\forall k}e^{\log(c^{k}) -\frac{1}{2}\sum_{\forall t}\left ( (x_{t}-\hat{x}_{t}^{k})^{2}+ (y_{t}-\hat{y}_{t}^{k})^{2} \right)- a^{*}}\\
		& \text{and} \\
		& a^{*} = \max_{\forall k} \left( \log(c^{k}) -\frac{1}{2}\sum_{\forall t}\left ( (x_{t}-\hat{x}_{t}^{k})^{2}+ (y_{t}-\hat{y}_{t}^{k})^{2} \right)\right)
	\end{split}
\end{align}
Here we use the log-sum-exp trick by subtracting the maximum value $a^{*}$ from each exponent, so that all exponentiations are of non-positive numbers and therefore overflow is avoided.

In addition, we believe that the use of domain knowledge may help improve the prediction accuracy of the model. For example, we prefer the predicted hypotheses to be in the drivable area of the scene. Therefore, we hope that the loss can help the model learn feasible trajectories, rather than infeasible ones. With this in mind, we explore the effect of introducing auxiliary off-road loss $L_{o}$, which helps to learn a more compliant trajectory relative to the drivable area. Given the detailed map $\mathcal{M}$, we can readily divide the map into multiple identical grids. The points of upper-left, upper-right, bottom-left, and bottom-right vertices of the grid $n$ is denoted as $n^{A}$, $n^{B}$, $n^{C}$, and $n^{D}$, respectively. Once the predicted way-point falls into a grid with no drivable roadways (we denote the set of this kind of grid as $N_{\bar{D}}$), the trajectory is considered to be infeasible and a penalty will be imposed. Formally, we consider the loss as
\begin{equation}\label{eq7}
	 L_{o}=\exp \left( \sum_{\forall k \in K, t\in (1:T), n \in N_{\bar{D}} }\eta[n\cap (\hat{x}_{t}^{k},\hat{y}_{t}^{k})]\right)
\end{equation}
where $\eta$ is a sign function, and is given as
\begin{equation}\label{eq8}
	\eta[n\cap (\hat{x}_{t}^{k},\hat{y}_{t}^{k})] = \left\{\begin{matrix}
		1 & \text{if}\ (\hat{x}_{t}^{k},\hat{y}_{t}^{k})\ \text{in}\ n\\
		0 & \text{otherwise}
	\end{matrix}\right.
\end{equation}
In order to determine the way-point $X= (\hat{x}_{t}^{k},\hat{y}_{t}^{k})$ is in the grid $n$ or not, we just need to calculate $\left( \overrightarrow{n^{A}n^{B}} \cdot \overrightarrow{n^{A}X} \right) \cdot \left( \overrightarrow{n^{C}n^{D}} \cdot \overrightarrow{n^{C}X} \right)$ and  $\left( \overrightarrow{n^{D}n^{A}} \cdot \overrightarrow{n^{D}X} \right) \cdot \left( \overrightarrow{n^{B}n^{C}} \cdot \overrightarrow{n^{B}X} \right)$. If these two terms are both larger than $0$, then the point $X$ is in $n$.

Since the total loss function can be viewed as the summation of distinct tasks, we use a multi-task learning approach to balance them,
\begin{equation}\label{eq9}
	L=\dfrac{1}{\sigma_{1}^{2}}L_{c} + \dfrac{1}{\sigma_{2}^{2}}L_{o}+\sum_{i=1,2}log(\sigma_{i}+1)
\end{equation}
\section{Experimental Analysis and Evaluations}
\subsection{Dataset}

\begin{figure}[h]
	\centering
	\includegraphics[width=1\linewidth]{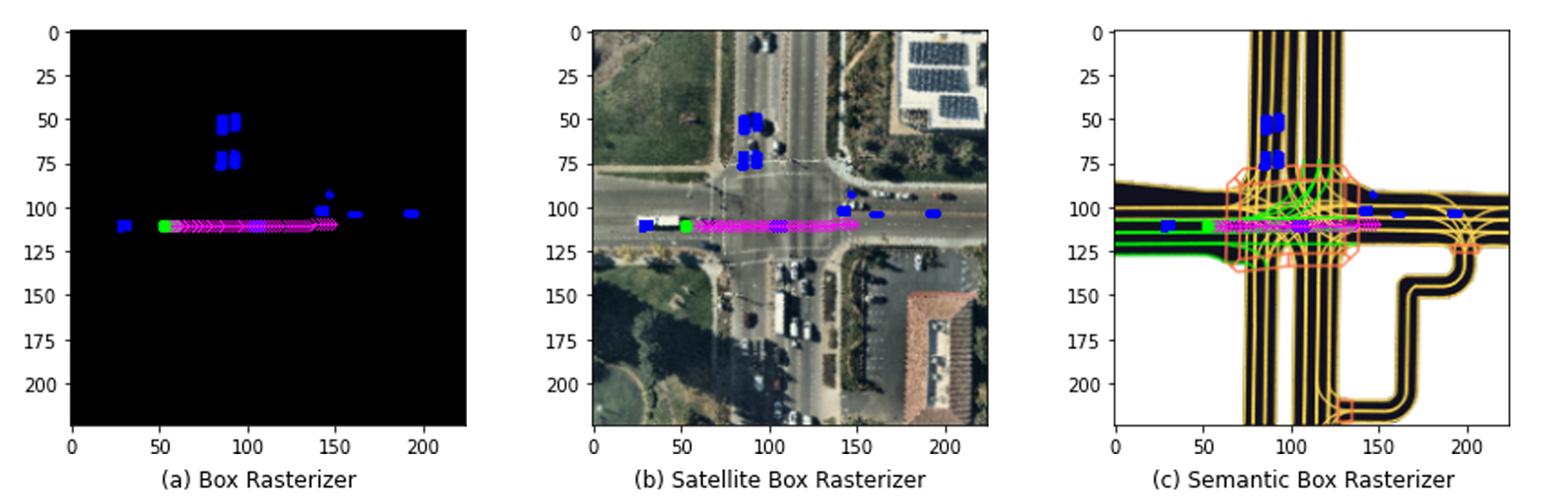}
	\caption{Examples of different BEV scene rasterization maps including AV (green rectangle) and TVs (blue rectangle). }
	\label{fig:basic-maps}
\end{figure}

\begin{figure*}[h]
	\centering
	\includegraphics[width=0.7\linewidth]{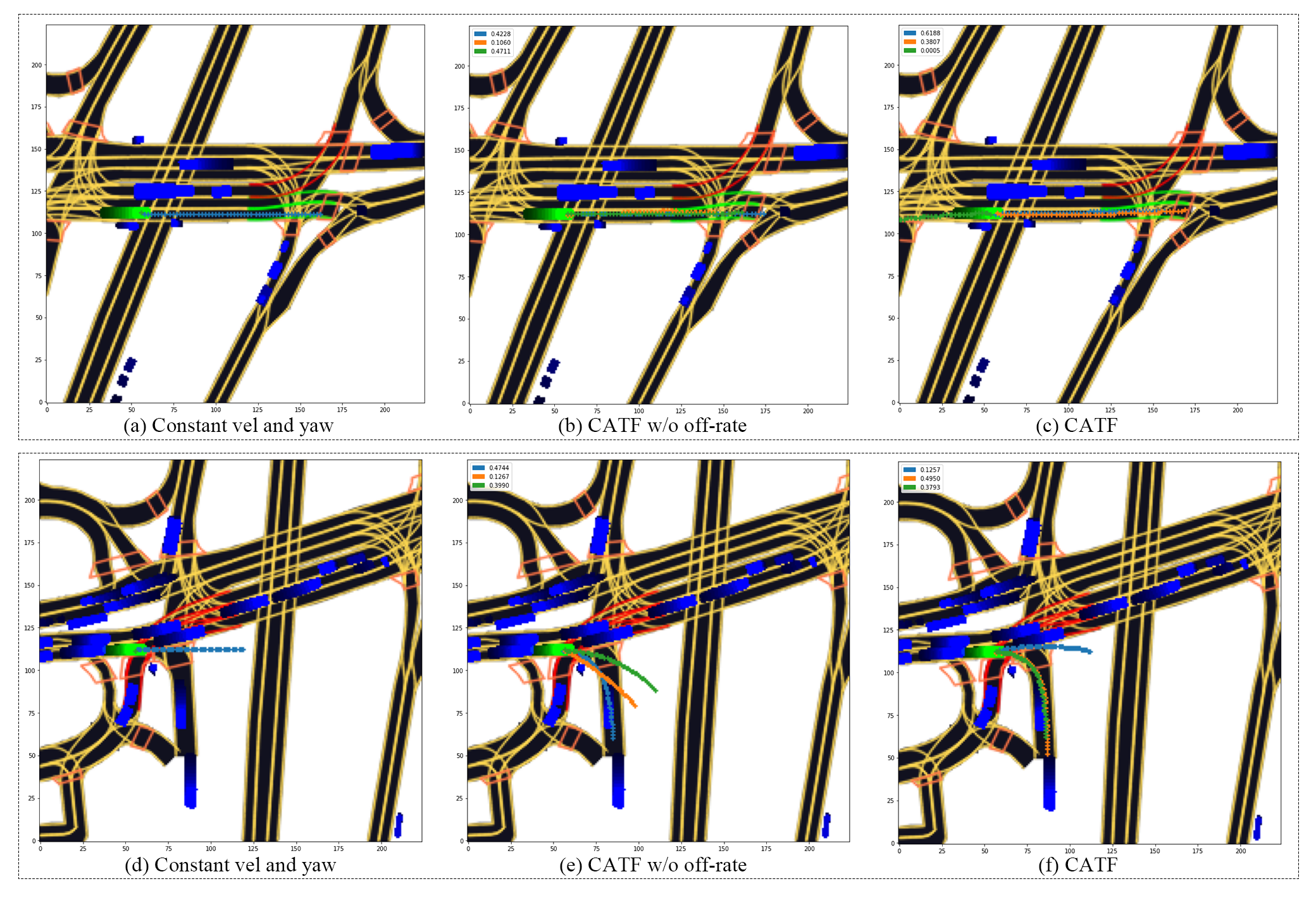}
	\caption{Model comparison in going straight (upper three) and turning scenes (lower three). Green rectangle represents the target agent, blue rectangle represents other agent,  darker afterimage indicate the history trajectory  ($K=3$, $h=1s$ and $H=5s$)}
	\label{fig:trajectory}
\end{figure*}
All the experiments in this paper are conducted on the Lyft autonomous driving dataset which is the largest and most detailed dataset available for training prediction and planning solutions as of now. It contains a total 1,118 hours of logs between October 2019 and March 2020 (approximately 162,000 scenes, each 25 seconds long), capturing the movement of autonomous vehicle, traffic participants around it and the status of the traffic lights. Fig. \ref{fig:basic-maps} presents an overview of a sample scene using different BEV rastersations. Since the ground truth is not provided in the original test set, we divide the original training set (134,000 scenes) into training, validation, and test sets, each of them contains around 70\%, 15\%, and 15\% of scenes in the entire set.
\subsection{Baselines}
We compare our work with the following prediction methods.
\paragraph{Constant Velocity and Yaw} ~\cite{scholler2020constant}: Observe historical positions of the agent and assumes the agent continues to move with the same velocity.
\paragraph{Multiple-Trajectory Prediction (MTP)} ~\cite{cui2019multimodal}: The MTP model uses a CNN over a rasterized representation of the scene and the agents to generate a fixed number of trajectories and their associated probabilities.
\paragraph{Trajectron} ~\cite{ivanovic2019trajectron}: The model combines variational deep generative models, LSTMs, and dynamic spatiotemporal graphical structures to produce multimodal trajectories.
\paragraph{ResNet-50} ~\cite{houston2020one}: Lyft official baseline. Using only semantic maps as input and a single ResNet-50 to predict future trajectories.

\subsection{Metrics}
We use commonly reported metrics to gain insight into various aspects of multi-modal trajectory prediction. In our experiment, we averaged the following metrics in all instances.
\paragraph{$\text{minADE}_{K}$ and $\text{minFDE}_{K}$:}We report average displacement error (ADE) and final displacement errors (FDE) over $K$ most probable trajectories. $\text{minADE}_{K}$ is

\begin{equation}\label{eq10}
	\min_{\forall k}\dfrac{1}{\mathcal{N}_{t_{j}}\cdot H}\sum_{i=1}^{N}\sum_{t= t_{j+1}}^{t_{j+H}}\parallel \mathbf{s}_{i,t} - \hat{\mathbf{s}}_{i,t}^{k}\parallel_{2}
\end{equation}
 and $\text{minFDE}_{k}$ is
\begin{equation}\label{eq11}
	\min_{\forall k}\dfrac{1}{N}\sum_{i=1}^{N}\parallel \mathbf{s}_{i,t_{j+H}} - \hat{\mathbf{s}}_{i,t_{j+H}}^{k}\parallel_{2}
\end{equation}

\paragraph{Off-road rate:} The off-road rate $r_{o,K}$ is used to measure the fraction of predicted trajectories that fall outside the drivable area of the map,
\begin{equation}\label{eq12}
	r_{o,K}=\dfrac{1}{K \cdot H \cdot \mathcal{N}_{t_{j}}} \sum_{k=1}^{K}\sum_{i=1}^{N}\sum_{t= t_{j+1}}^{t_{j+H}}\eta[n\cap (\hat{\mathbf{s}}_{i,t}^{k})] |_{\forall n \in N_{\bar{D}}}
\end{equation}

\begin{table*}[h]
	\centering
	\begin{tabular}{@{}llllllll@{}}
		\toprule
		& $\text{minADE}_{1}$            & $\text{minADE}_{3}$            & $\text{minADE}_{6}$            & $\text{minFDE}_{1}$            & $\text{minFDE}_{3}$            & $\text{minFDE}_{6}$            & $r_{o,3}$                      \\ \midrule
		Constant vel and yaw & 4.72                           & 4.72                           & 4.72                           & 12.03                          & 12.03                          & 12.03                          & 0.19                           \\
		MTP                  & 4.24                           & 2.55                           & 1.96                           & 10.24                          & 4.52                           & 3.12                           & 0.14                           \\
		Trajectron             & 3.72                       & 2.02                           & 1.88                           &9.22                          & 3.64                         & 2.66                          & 0.16                           \\
		ResNet-50            & 3.88                           & \textbf{1.88} & 1.72                           & 8.65                           & 3.54                           & 2.81                           & 0.18                           \\
		TF                   & 3.98                           & 2.12                           & 1.69                           & 9.23                           & 3.68                           & 2.52                           & 0.13                          \\
		TF w/ off--road         & 4.02                           & 2.14                          & 1.70                           & 9.12                           & 3.62                           & 2.55                           & \textbf{0.07 }                         \\
		CATF w/o off-road    & 3.81                           & 2.06                           & \textbf{1.63} & \textbf{8.35} & 3.52                           & 2.31                           & 0.13                           \\
		CATF                 & \textbf{3.66} & 1.89                           & 1.64                           & 8.41                           & \textbf{3.34} & \textbf{2.25} & \textbf{0.07} \\
		$\text{CATF}_{l}$    & 3.67                         & 1.91                          & 1.64                          & 8.41                           & 3.35                          & \textbf{2.25} & \textbf{0.07} \\ \bottomrule
	\end{tabular}
	\caption{Results of comparative analysis with $h=1s$ and $H=5s$}
	\label{tab:my-table}
\end{table*}

\subsection{Implementation Details}
For the CATF model, we adopt the same parameters of the original Transformer Networks, namely $D= 512$, $N=6$, $h=8$, and $d_{k}=64$. For the $\text{CATF}_{l}$ model, $p$ is set to be 64. The input raster is the RGB image of size $(224\times224\times3)$. The origin is set to be at bottom left, and the resolution is 0.25 meters per pixel. Both models use a MobileNet-V3 backbone with pretrained ImageNet weights to extract map features.  In addition, the masked technique is also implemented in the decoder to prevent positions from attending to subsequent positions. The batch size is 64. We train the networks via backpropagation with the Adam optimizer, linear warm-up phase for the first 5 epoch and a decaying learning rate afterward with dropout value of 0.1.

\subsection{Ablation Study and Quantitative Results}
We compare the CATF and $\text{CATF}_{l}$ to various baselines in Table \ref{tab:my-table}. The TF model is the vanilla Transformer model and without context input and off-road penalty,  and TF w/off-road model indicates the off-road loss is added when training the model. It can be seen that all the TF-based models have better performance than LSTM and CNN based baselines, indicating the Transformer model has better abilities in dealing with this kind of sequence data. Our CATF model outperforms all baselines on 4 of the 7 reported metrics, while being second on the remaining three, representing the sate of the art on the l5kit benchmark as of writing this paper.  The $\text{CATF}_{l}$ also has top-tier performance among all methods while it has about $1.6\times$ faster inference time and allows for a $1.8\times$ smaller memory usage than CATF. These results show that the introduction of linear projection technology can notably reduce the computational burden and without having significant degradation on the results.

Moreover, our methods present improvements compared to others considering the off-road rate metric, especially when trained with the off-road loss that penalizes predictions outside of the drivable area. The results show that the proposed methods can generate trajectories that conform to the scene. Fig. \ref{fig:trajectory} presented detailed model comparison results in both going straight and turning scenes. It can be seen that since multi-modal models can predict multiple trajectories with associated confidences, they are more likely to generate feasible trajectories than uni-modal models (e.g., physics-based models). As shown in Fig. \ref{fig:trajectory}, in the turning scene, this drawback becomes more significant. The comparison between Fig. \ref{fig:trajectory}(e) and (f) shows the importance of introducing off-road loss very intuitively. In Fig. \ref{fig:trajectory}(e), only one trajectory successfully predicted the turning process of the vehicle with a probability of 0.47. While in the prediction result of the CATF model, the probability is significantly increased to 0.87.


\section{Concluding Remarks}
In this work, we have introduced a Context-Aware Transformer model for the task of surrounding agent's trajectory prediction for autonomous driving. The model is able to utilize the information of both contextual cues and historical states of agents. A novel off-road loss is also introduced to constraint predictions to be feasible. The proposed model outperforms baselines with a large margin in experiments conducted on the Lyft l5kit dataset.


\bibliographystyle{named}
\bibliography{v1ijcai21li}

\end{document}